\documentclass[11pt]{article}
\usepackage[letterpaper]{geometry}

\usepackage{times}
\usepackage{latexsym}
\usepackage{url}
\usepackage{amsmath}
\usepackage{amsfonts}
\usepackage{booktabs}
\usepackage{multirow}
\usepackage[T1]{fontenc}
\usepackage[utf8]{inputenc}
\usepackage{graphicx}
\usepackage{caption}
\usepackage{placeins}
\usepackage{booktabs}
\usepackage{siunitx} 
\usepackage{longtable}
\usepackage{acl} 
\usepackage{lineno}
\usepackage[inline]{enumitem}

\title{Improving Handshape Representations for Sign Language Processing: \\ A Graph Neural Network Approach}

\author{
  Alessa Carbo \\
  Johns Hopkins University \\
  \texttt{acarbol1@jh.edu} \And
  Eric Nalisnick \\
  Johns Hopkins University \\
  \texttt{nalisnick@jhu.edu}
}

\date{}

\begin{document}
\maketitle

\begin{abstract}
Handshapes serve a fundamental phonological role in signed languages, with American Sign Language employing approximately $50$ distinct shapes.  However, computational approaches rarely model handshapes explicitly, limiting both recognition accuracy and linguistic analysis. We introduce a novel graph neural network that separates temporal dynamics from static handshape configurations.  Our approach combines anatomically-informed graph structures with contrastive learning to address key challenges in handshape recognition, including subtle inter-class distinctions and temporal variations. We establish the first benchmark for structured handshape recognition in signing sequences, achieving 46\% accuracy across 37 handshape classes (with baseline methods achieving 25\%). 
\end{abstract}

\section{Introduction}
Sign languages are sophisticated linguistic systems in which meaning emerges from multiple parameters: handshapes, movement, location, palm orientation, and nonmanual markers \citep{sandler2006sign}. Handshapes serve as fundamental phonological units.  For example, \textit{American Sign Language} (ASL) employs approximately 40 to 50 distinct handshapes that can create minimal pairs: pair of signs that are identical in all but one of the five phonological parameters \citep{stokoe2004sign, brentari2011handshape}.

Despite their linguistic importance, computational approaches rarely model handshapes explicitly. Existing systems for sign language recognition process a signing sequence holistically \citep{shi2022open, dilsizian2014framework}, leaving handshape information implicit, which limits both recognition accuracy and linguistic analysis \citep{huenerfauth2009sign}. This gap stems from limited phonological annotations \citep{caselli2016asl} and the challenge of modeling both discrete linguistic categories and continuous physical variations \citep{brentari2011handshape}.  Yet \citet{zhang2023handshape} show that explicitly modeling handshapes benefits downstream processing tasks, such as improving translation accuracy by 15\%. 
 \citet{koller2015continuous} find that phonological features enhanced recognition robustness across signers.

In this work, we propose a novel \textit{graph neural network} (GNN) that explicitly separates temporal dynamics from static handshape configurations in continuous signing sequences. A key challenge in handshape recognition is that the signer's hand configuration evolves dynamically throughout the signing sequence. This temporal-static tension motivates our dual GNN architecture: one sub-model captures the temporal evolution of hand configurations throughout the sign, and another sub-model focuses on representative static frames in which the handshape can be recognized in a canonical form.  Given the absence of prior baselines for this task, our work establishes the first benchmark for handshape recognition as a standalone task within sign language processing (SLP).

\section{Related Work}

\paragraph{Modeling Handshapes} Modeling hands is a long-studied problem in computer vision, with recent advancements focusing on end-to-end deep learning methods to predict hand shape and pose from RGB images 
\citep{oudah2020survey}.Previous work on recognizing handshapes for sign language has primarily treated handshape recognition as only one component of a broader phonological modeling framework. 

In work that is perhaps the most related, \citet{kezar2023exploring} explored multi-task and curriculum learning strategies for recognizing sixteen phoneme types, including handshapes. 
 Their approach differs from ours by focusing on a multi-task and curriculum learning setting that requires extensive phonological annotations ($16$ per sign) and relies on substantial pretraining.  They use a pretrained sign recognition model followed by a supervised training step involving sign glosses.~Yet \citet{kezar2023exploring} also use a GNN---specifically, the graphical convolutional NN proposed by \citet{jiang2021skeleton}.  This architecture uses a key-point skeletal representation of the hands, arms, and neck.  We, on the other hand, propose a GNN architecture tailored specifically to handshape recognition, without pre-training or extensive annotation requirements, which, ostensibly, could be integrated into these broader pipelines. 
 

\paragraph{Incorporating Handshape Information in SLP}
\citet{zhang2023handshape} demonstrated that explicit handshape modeling improves the performance of SLP translation systems by augmenting the PHOENIX14T dataset with handshape labels (PHOENIX14T-HS). Similarly, \citet{tavella2022wlasl} developed WLASL-LEX, annotating comprehensive phonological properties including handshapes, motion, and location, emphasizing the importance of phonological features in datasets for SLP.

\paragraph{Graph Neural Networks}
GNNs have shown promise in capturing spatial-temporal relationships in skeletal data. Building on foundational work such as GCN-BERT \citep{tunga2021pose} and ST-GCN, architectures such as HST-GNN \citep{kan2022hierarchical} and SL-GCN \citep{jiang2021skeleton} demonstrate the utility of modeling hierarchical structures for SLP.

\paragraph{Relationship to our Work}
Prior sign language recognition models using CNNs or RNNs focus on full signing sequences rather than explicit handshape modeling \citep{shi2022open}.  Our work advances these efforts through a dual-network GNN architecture that separately addresses temporal and static aspects of handshape configuration. Our GNN employs a hierarchical structure differentiating ourselves from previous approaches that process sign language uniformly, that is, with the same graph structure throughout. 
\citep{koller2015continuous}.

\section{A Graph Neural Network for Handshape Representations}
We now introduce our novel GNN for handshape recognition, which we call \textit{Handshape-GNN}. 
 In sign language corpora, handshapes evolve temporally over multiple frames, often taking their canonical form is only a subset of these frames.  The other frames typically show the transitions between handshapes. Thus a robust model for handshape recognition must be aware of these handshape dynamics. Our Handshape-GNN explicitly separates temporal dynamics from static configurations by having two specialized sub-networks: one captures the temporal evolution of the hands throughout a complete sign, which will often contain multiple handshapes.  The second component analyzes static frames, searching for the handshape in its canonical, and therefore most recognizable, form.~GNNs are particularly well-suited for this task: they can process structured data with irregular or otherwise complex relationships---a characteristic that has made them highly effective in domains ranging from social networks to protein folding  \citep{kipf2017semi,xu2019how}.

\subsection{Handshape Data}
Modern hand tracking systems such as \texttt{MediaPipe} represent hands as a set of $21$ keypoints (joints and fingertips).  This representation naturally becomes a graph by considering the keypoints as nodes and with the hand anatomy defining the edges.  The data representing one or more signs takes the form of a sequence $S = \{\mathbf{X}_1, \mathbf{X}_2, \ldots, \mathbf{X}_T \}$.  Each $\mathbf{X}_t \in \mathbb{R}^{21 \times 3}$ represents the three dimensional coordinates of the $21$ hand landmarks at time $i$. Each sequence has a corresponding handshape label set, which denotes the one or more handshapes that are required to produce the sign(s): $Y = \{y_1 , \ldots, y_j \}$, where $y \in [1, K]$, with $K$ denoting the total number of possible handshapes.  Our full data set is then comprised of $N$ sequences: $\mathcal{D} = \{ (S_n , Y_n ) \}_{n=1}^{N}$.  In our experiments, we use the PopSign dataset, meaning that $S_n$ represents an isolated sign (as detailed in Section 4.2).~We leave the application of our approach to datasets with multiple signs per sequence to future work.  




\begin{figure}[t]
\centering
\includegraphics[width=0.40\textwidth]{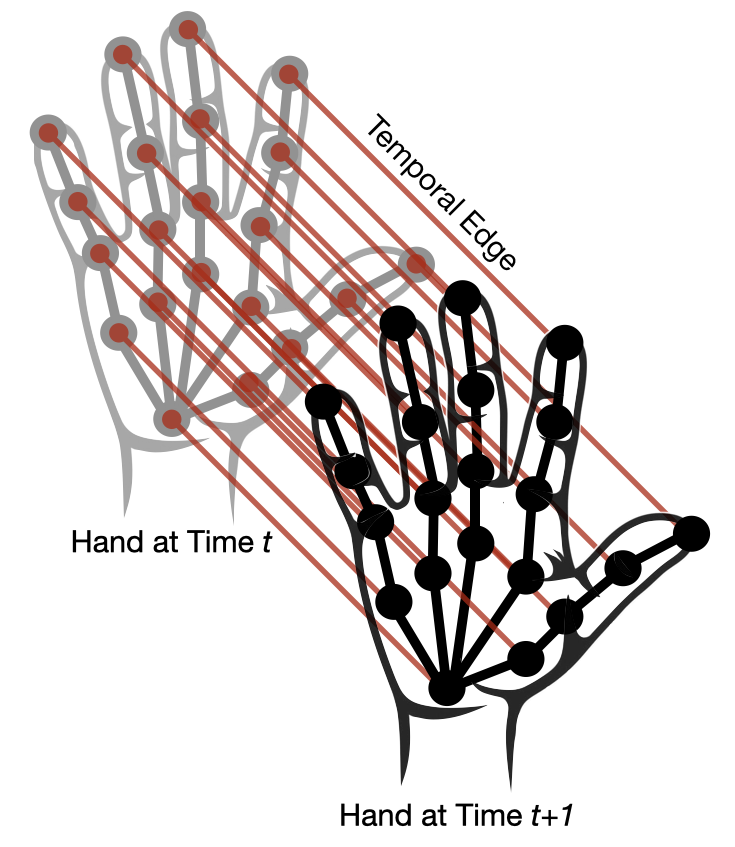}
\caption{\textit{Spatial-Temporal Keypoints on Hand}. This visualization shows the 21 MediaPipe keypoints that form the spatial representation of the hand.  Anatomical edges are shown in black, temporal edges (across frames) are shown in red.}
\label{fig:hand_keypoints}
\end{figure}
\subsection{Sub-Model \#1: Sign GNN}
~The first component of our dual architecture is the sign GNN.  It processes the full frame sequence through a graph structure that captures both spatial and temporal relationships. The network updates node features through message passing, with each node (keypoint) aggregating information from its neighbors. To represent the temporal and spatial relationships, the graph in our GNN is constructed with two types of edges:
\begin{itemize}
    \item Spatial edges $E_s$ connecting landmarks within each frame: $E_s = \{(i,j) \mid i,j \text{ are anatomically connected}\}$.  These represent physical connections such as finger joints.
    \item Temporal edges $E_t$ connecting each landmark to itself across consecutive frames: $E_t = \{(i_t, i_{t+1}) \mid i \text{ is a landmark index}\}$.  This captures how each point moves through time.
\end{itemize}
See Figure \ref{fig:hand_keypoints} for a visualization of both the spatial (black) and temporal (red) edges.

This architecture processes information hierarchically, mirroring the natural structure of sign language. At the lowest level, it encodes individual keypoint positions. These features are then aggregated to capture finger-level patterns and coordinated movements. Finally, at the highest level, the network learns complete hand configurations and their temporal evolution over the signing sequence.

\paragraph{Architecture} For each layer $l$, the node features $\mathbf{H}_l$ are updated according to:
\[\mathbf{H}_{l+1} = \text{\texttt{LeakyRELU}}\left(\mathbf{D}^{-1/2}  \tilde{\mathbf{A}} \mathbf{D}^{-1/2} \mathbf{H}_{l} \mathbf{W}_{l} \right)\]
where $\tilde{\mathbf{A}} = \mathbf{A} + \mathbb{I} $, which represents the connections between landmarks (including self-connections via $\mathbb{I}$), $\mathbf{D}$ is the corresponding degree matrix $(21T \times 21T)$ that normalizes message passing based on node connectivity. $\mathbf{W}_{l}$ are learnable weight matrices with dimensions $(d_{l} \times d_{(l+1)})$, where $d_{l}$ is the input feature dimension and $d_{(l+1)}$ is the output dimension for layer $l$. Our implementation stacks three of these layers, transforming the initial 3-dimensional position features through hidden dimensions of 64 and finally to 32 features per node ($3\rightarrow64\rightarrow64\rightarrow32$). Between layers, we apply batch normalization to stabilize training and dropout to prevent overfitting.


\paragraph{Loss Function} We train this network using contrastive learning  \citep{vandenoord2018contrastive} with a binary cross-entropy objective.~For a batch of signing sequences, we compute: \begin{equation*}
    \begin{split}
      \ell&\left(\theta; \mathbf{B}\right) = 
-\frac{1}{|\mathcal{P}|} \sum_{(i,j)\in \mathcal{P}} \log \sigma\left(s^{\theta}_{ij}\right) \\
 & \quad \quad \quad \quad   -\frac{1}{|\mathcal{N}|} \sum_{(i,j)\in \mathcal{N}} \log \left(1 - \sigma\left(s^{\theta}_{ij}\right)\right)  
    \end{split}
\end{equation*} where $\theta$ denotes all model parameters and $B$ is a batch of sequences with embeddings $\{\mathbf{z}_{1}, \mathbf{z}_{2}, \dots, \mathbf{z}_{B}\}$.  $\mathbf{z}$ represents the final graph-level embedding obtained by pooling node features from the last GCN layer $\mathbf{H}_{L}$ across all nodes in sequence $S_n$.  $\sigma(\cdot)$ is the sigmoid (logistic) function, and $s^{\theta}_{ij} = \tau \cdot \mathbf{z}_i^{\top} \mathbf{z}_j / \|\mathbf{z}_i\| \cdot \|\mathbf{z}_j\|$ is the cosine similarity scaled by temperature $\tau \in \mathbb{R}^{+}$.  $\mathcal{P} = \{(i,j) \mid y_i = y_j, i \neq j\}$ are positive pairs (same sign label), and  
$N = \{(i,j) \mid y_i \neq y_j, i \neq j\}$ are negative pairs (different sign labels).

Our use of sign-level label supervision may at first seem orthogonal to our ultimate goal of handshape recognition. Yet, since our sequential data lacks frame-level handshape annotations, we leverage the fact that similar temporal dynamics often employ related handshape transitions and movement patterns, which provides a useful proxy signal. Additionally, this approach helps address the limitation of having only a monocular view, which causes handshape details to be partially obscured. 

\subsection{Sub-Model \#2: Handshape GNN}
While the Sign GNN captures temporal dynamics, our second sub-model provides a complementary, static analysis of handshapes. This Handshape GNN focuses on representative static frames in which hand motion is minimal and all keypoints are reliably detected.

\paragraph{Identifying Candidate Frames}  Rather than attempting to manually annotate handshape timestamps---a laborious, somewhat subjective task---we use motion-based heuristics to select representative frames. We identify low-motion frames by computing average keypoint displacement in adjacent frames: $
v_t \ = \ ||\mathbf{X}_{t} - \mathbf{X}_{t+1} ||_{2}$. In practice, we select the single frame with the minimum $v_t$ value per signing sequence. While our frame selection heuristic may not always identify the optimal handshape view, this is hopefully made up for by having the additional temporal representation. 

\paragraph{Architecture and Graph Structure} The Handshape GNN uses the same graph convolutional architecture and message passing operation as the Sign GNN, but it operates exclusively on spatial connections within a single frame.   While maintaining the same fundamental structure of 21 nodes per frame, our implementation features an anatomically-informed bidirectional edge structure:
\begin{tabular}{@{}p{0.3\linewidth}p{0.65\linewidth}@{}}
\textbf{Sequential:} & 
$\{(j_k, j_{k+1}), (j_{k+1}, j_k) \mid j_k \text{ is the } k\text{-th joint of a finger}\}$ \\[2pt]
\textbf{Cross-finger:} &
$\{(j_k^i, j_k^{i+1}), (j_k^{i+1}, j_k^i) \mid j_k^i \text{ is joint } k \text{ of finger } i\}$ \\[2pt]
\textbf{Palm-centered:} &
$\{(w, b_i), (b_i, w) \mid w \text{ is wrist, } b_i \text{ is finger base } i\}$ \\[2pt]
\textbf{Diagonal palm:} &
Linking the thumb base with finger bases
\end{tabular}


Like the SignGNN, this network processes information hierarchically through its three convolutional layers from single keypoint representation to the full hand configuration.  The dimensions similarly match the architecture of the Sign GNN.  We train this network using the same contrastive learning with a binary cross-entropy objective as the SignGNN, but now pairs are defined by their handshape label (instead of their sign label).


\subsection{Combined Classification Framework}
\begin{figure}[t]
\centering
\includegraphics[width=0.5\textwidth]{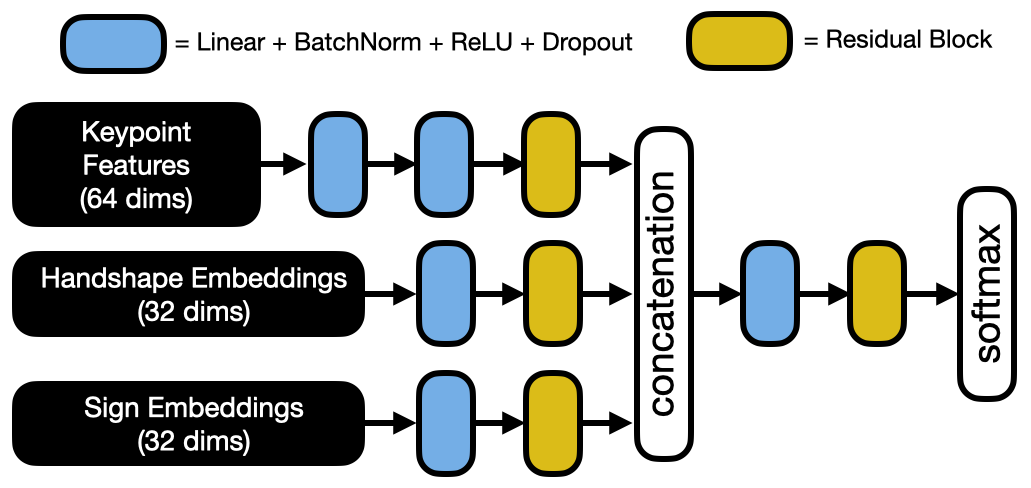}
\caption{\textit{Full Model Architecture.} Above is a diagram of our triple-stream architecture that combines SignGNN embeddings, HandshapeGNN embeddings, and raw landmarks to produce a final handshape classification.}
\label{fig:architecture}
\end{figure}

To leverage the complementary strengths of the two sub-models, we develop a triple-stream classification architecture that combines temporal dynamics, static configurations, and raw geometric information.  See Figure \ref{fig:architecture} for a diagram of the full model architecture.  Our framework processes three parallel input streams: sign embeddings (32 dimensional) from the Sign GNN, handshape embeddings (32 dimensional) from the Handshape GNN, and raw landmarks (63 dimensional) capturing direct geometric information.~This multi-stream design allows us to leverage both the specialized representations learned by each GNN while preserving direct access to the geometric features, ensuring our model can capture both high-level patterns and fine-grained details.

Each component first processes its input separately. The baseline classifier processes raw landmark coordinates through a three-layer network (63→256→256→37) with batch normalization and dropout ($p = 0.3$). For the GNN components, we extract pre-trained embeddings and train separate classifiers. SignGNN, we trained a classifier that processes 32-dimensional sign embeddings through residual blocks with batch normalization and dropout. The HandshapeGNN classifier similarly transforms 32-dimensional handshape embeddings through multiple residual blocks.  

In our final combined architecture, each stream's embeddings are passed through dedicated processing layers (32→64 dimensions) with batch normalization and dropout ($p = 0.2$). These features are then concatenated and passed through a final classification layer using standard cross-entropy loss over the 37 handshape classes. This multi-stream approach allows the classifier to leverage both the learned representations from our GNN networks and the raw geometric information, providing a robust framework for handshape recognition.   

\begin{figure}[t]
\centering
\includegraphics[width=0.45\textwidth]{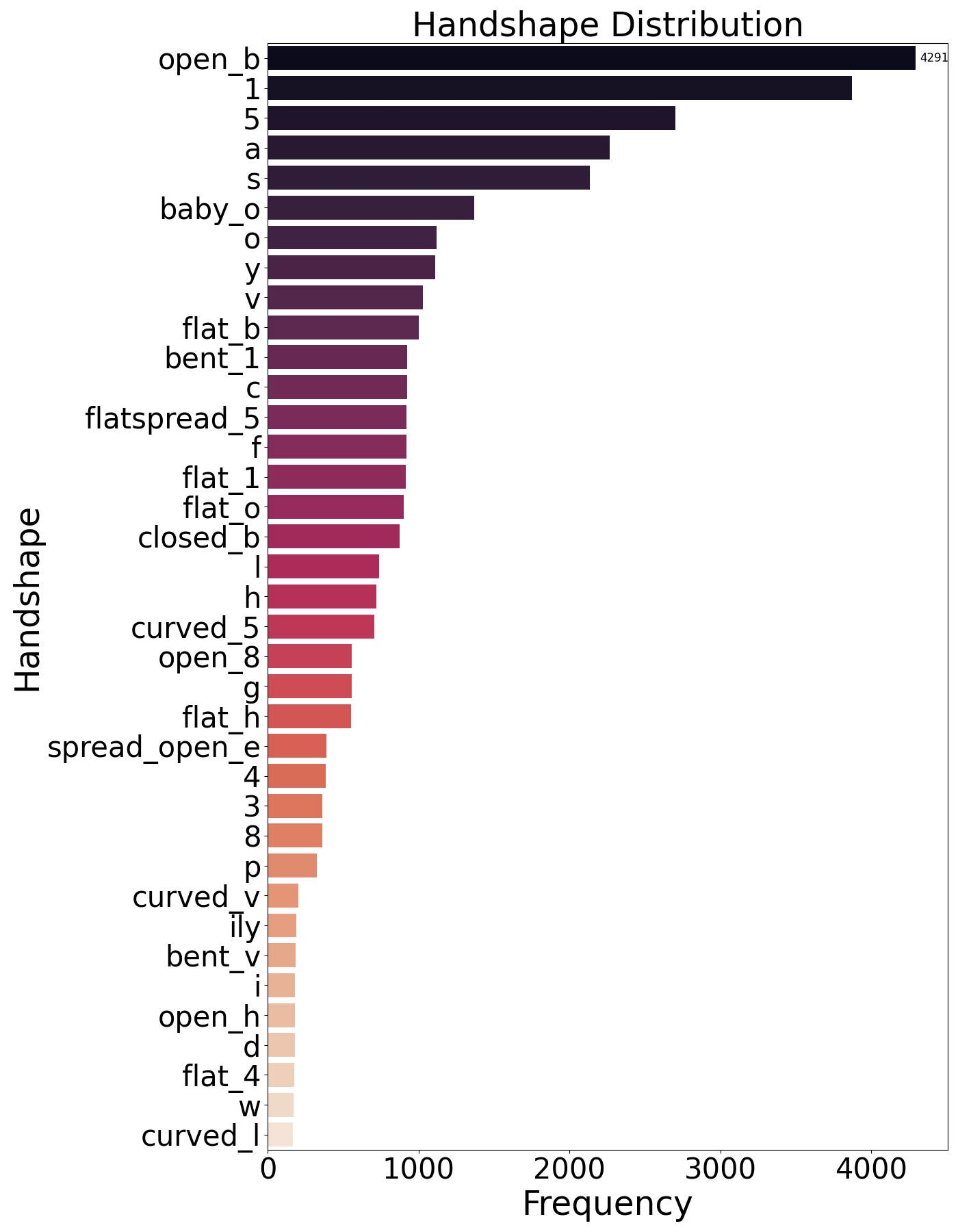}
\caption{\textit{Histogram of Handshape Classes.} Distribution of handshapes in our dataset, which shows extreme class imbalance. }
\label{fig:handshape_distribution}
\end{figure}
\section{Dataset Construction}\label{sec:dataset}
We create a dataset that supports sign-language-based handshape recognition by combining \texttt{PopSign} \citep{starner2023popsign}, which provides keypoint sequences for over 200,000 instances of isolated signs, with \texttt{ASL-LEX} \citep{caselli2016asl}, a lexical database with phonological annotations for over 2,700 ASL signs. We selected PopSign as it provides data already processed into MediaPipe landmark format,  and though we limited this analysis to this dataset in order to maintain consistency and focus on the architecture proposed, there are other isolated sign datasets that could have been selected. The resulting 34,533 samples (27,626 training, 6,907 validation) represent $37$ distinct handshapes based on the Prosodic model \citep{brentari2011handshape}, each constructed as graph representations preserving spatial and temporal relationships.  Most frequent are \textit{open b} (4,291 instances), \textit{1} (3,872 instances), and \textit{5} (2,699 instances). Figure \ref{fig:handshape_distribution} shows a histogram of the handshape frequencies, which is long-tailed, as is expected.



\section{Metrics for Handshape Analysis}
We incorporate a set of biomechanical metrics adapted from \citet{yin2024american} that quantify handshape production effort and perceptual distinctiveness. These metrics quantify to what degree handshapes are influenced by both phonological distinctions and biomechanical constraints, potentially helping us understand why certain handshapes are easier to recognize than others.

\paragraph{Finger Independence} To calculate finger independence, we organize the hand landmarks into three functional groups: metacarpophalangeal joints (where fingers meet palm), proximal interphalangeal joints (middle knuckles), and distal interphalangeal joints (fingertip knuckles). The finger independence score for each group is:
\[\text{FI}(\mathbf{X}) = \sum_{J \in G_J} \sum_{i,j \in J, i<j} |\alpha_i - \alpha_j|\]
where $G_J$ is the set of joint groups, and $\alpha_i, \alpha_j$ are the joint angles for fingers $i$ and $j$, with higher scores indicating greater independence. This is calculated for each frame $\mathbf{X}$ (all 21 hand landmarks).

\paragraph{Thumb Effort} We compare joint angles to reference positions to quantify how relaxed the hand's thumb is, which corresponds to the effort required to make the handshape:
\[\text{TE}(\mathbf{X}) = \min_{r \in R} \frac{\sum_{i \in \mathcal{T}} |\alpha_i - \beta_i|}{N}\]
where $\mathbf{X}$ represents a single frame of hand landmarks, $\mathcal{T}$ is the set of thumb joints, $\beta_i$ is the corresponding joint angle in resting position $r$, $R$ is the set of reference resting hand configurations, and $N$ is the number of thumb joints. 

\paragraph{Handshape Distance Metric} To quantify perceptual distinctiveness between two hand configurations $\mathbf{X}_1$ and $\mathbf{X}_2$, we calculate:
\[D(\mathbf{X}_1, \mathbf{X}_2) = \frac{\sum_i |\alpha_i - \beta_i|}{N}\]
where $\alpha_i$ and $\beta_i$ are the joint angles in hand configurations $\mathbf{X}_1$ and $\mathbf{X}_2$ respectively, and $N$ is the total number of joints compared. This score provides a fair comparison of handshapes regardless of the number of joints involved.

\section{Experimental Results}
We now report our experimental findings on the \texttt{PopSign} dataset augmented with \texttt{ASL-Lex} handshape annotations, as was introduced in Section \ref{sec:dataset}.  We first report handshape classification accuracy and F1 score.~We then probe those results using the aforementioned handshape metrics, attempting to identify our and the baseline models' strengths and weaknesses.  We report the mean across 4 random seeds for all experiments.

\begin{table}[t]
    \setlength{\tabcolsep}{3pt}
    \begin{center}
    \caption{\textit{Accuracy and F1 Score on Handshape-Augmented PopSign.} The table below reports the accuracy and F1 scores of our dual approach against using only the constituent models (i.e.~sign-specific and handshape-specific GNNs) and a basic multilayer perceptron (MLP).}
    \label{tab:model-comparison}
    \begin{tabular}{@{}l@{\hspace{4pt}}c@{\hspace{4pt}}c@{}}
        \toprule
       Model & Accuracy (\%) & F1 Score \\
      \midrule
        MLP & 25.40 & 0.24 \\
        Sign GNN & 30.01 & 0.26 \\
        Handshape GNN & 31.00 & 0.26 \\
        \textbf{Dual GNN} & \textbf{46.07} & \textbf{0.44} \\
        \bottomrule
    \end{tabular}
    \end{center}
\end{table}

\subsection{Classification Results}
\paragraph{Hyperparameter Selection} For all models, we perform hyperparameter tuning using grid search over learning rates 
($1 \times 10^{-6}$ to $1 \times 10^{-3}$), weight decay values 
($1 \times 10^{-5}$ to $1 \times 10^{-4}$), and early stopping patience 
(2 to 75 epochs).~The best performing configuration used learning rate 
$1 \times 10^{-4}$, weight decay $1 \times 10^{-4}$, and patience of 50 epochs.

\paragraph{Accuracy and F1} The accuracy and F1 scores are reported in Table \ref{tab:model-comparison}.  Our dual GNN approach achieved significant improvements over baseline methods in handshape recognition. The baseline feedforward multilayer perceptron (MLP) achieved 25\% accuracy on raw landmark features, with single GNN variants reaching 30\% and 31\% accuracy through spatial and static modeling respectively. Our final combined architecture achieved 46\% accuracy across the 37 handshape classes.  This roughly 15\% improvement over the constituent sub-models suggests that they are encoding complementary information.

\begin{figure}[t]
\centering
\includegraphics[width=0.5\textwidth]{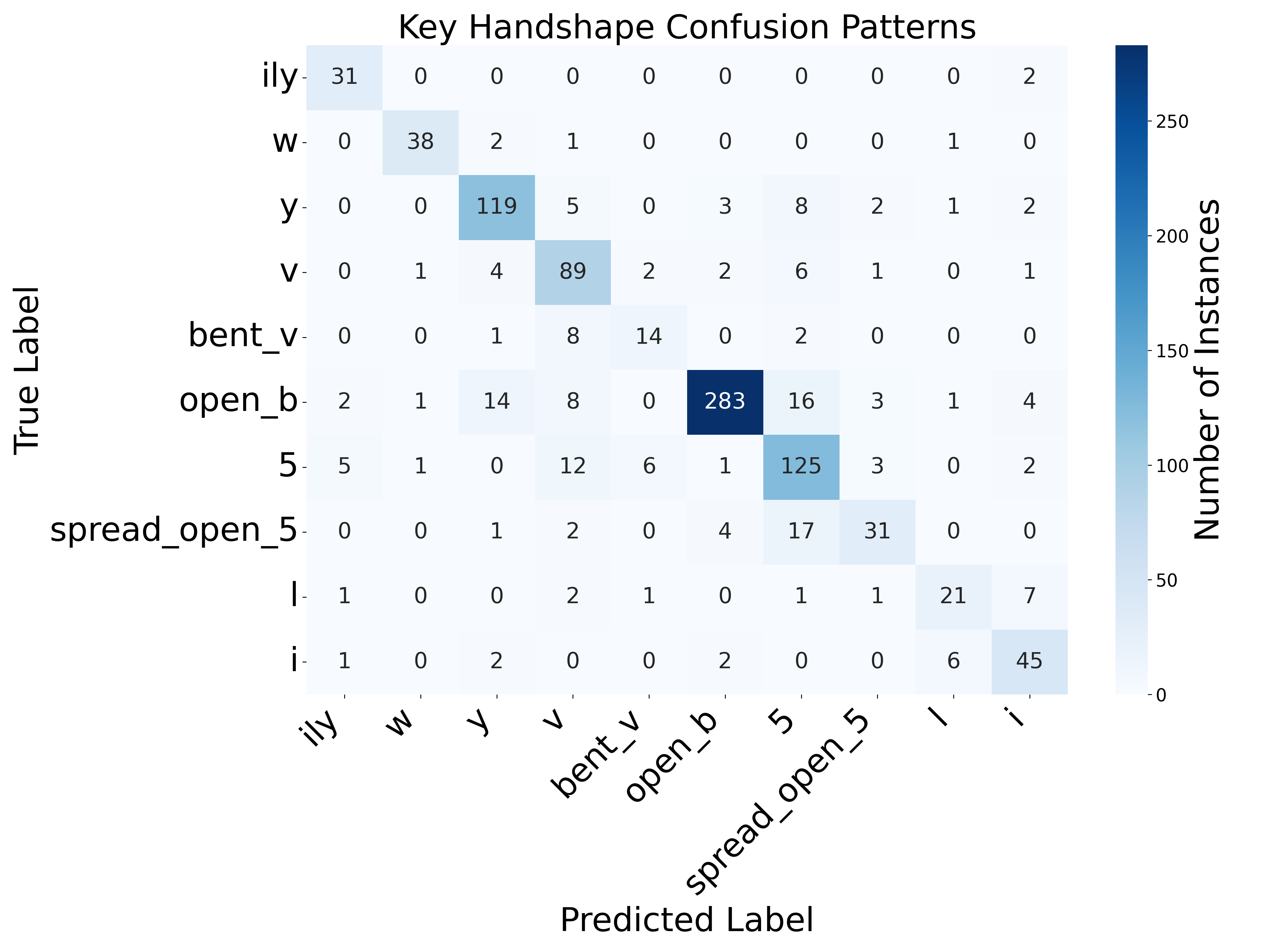} 
\caption{\textit{Confusion Matrix of Selected Classes.}  Above we show a selected confusion matrix for our dual GNN model. Errors often occurred among \textit{v}-variations and \textit{5}-variations.}
\label{fig:confusion_matrix}
\end{figure}

\begin{figure}[t]
\centering
\includegraphics[scale=0.235]{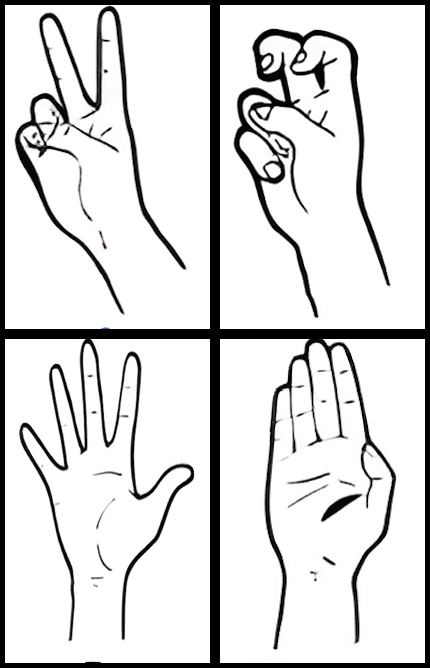}
\caption{\textit{Commonly Confused Handshapes.} Examples of visually similar handshapes that were confused by our model: \textit{v} (top left) vs \textit{bent v} (top right), \textit{5} (bottom left) vs \textit{flat b} / \textit{closed 5} (bottom right).}
\label{fig:handshape_examples}
\end{figure}

\paragraph{Error Analysis} See Figure \ref{fig:confusion_matrix} for a confusion matrix of selected classes.  Misclassifications primarily occurred between geometrically similar handshapes, especially within the \textit{5}-family due to their similar shapes involving finger spreading. Fine-grained joint angle differences also proved challenging (e.g., \textit{bent 5} vs \textit{claw 5}), highlighting the difficulty of distinguishing subtle variations in articulation.  We show these often confused handshapes in Figure \ref{fig:handshape_examples}.

The impact of class imbalance varied, with common classes such as \textit{open b} showing good recall but low precision, while some uncommon classes like \textit{ily} achieved strong performance due to its distinctiveness (F1 score: 0.73).  The dual GNN showed clear improvements over baselines in distinguishing handshapes that differ in palm orientation or in temporal evolution.  Handshapes with similar thumb positions showed notable confusion in baseline models.  A full breakdown of per-class results is provided in Appendix A2.


\subsection{Handshape Metrics Results}
Analysis of the biomechanical metrics revealed three key patterns (more details in Appendix A.3): a bimodal distribution in thumb effort scores, which reinforces the importance of the thumb configuration as a key feature for handshape classification; a strong mode at zero for finger independence, indicating that strong independence may be a helpful feature (e.g.~\textit{ily}); and a bimodal distribution in handshape distances. These patterns provide evidence that physical constraints lead to fundamental difficulties for handshape classification.


\section{Discussion and Conclusions}
Our dual GNN architecture demonstrates that incorporating both dynamic and static hand configurations is useful for handshape recognition, obtaining a 15\% accuracy improvement over methods that leverage only one modality.~Built from anatomically-informed graph structures and contrastive learning, our model bridges purely data-driven and linguistically-motivated approaches.~Importantly, our approach achieves these results without external pretraining dependencies or large-scale data requirements.  While we have provided a strong initial model for isolated handshape recognition, challenges remain due to the geometric similarly within some handshape subfamilies.




\section{Limitations and Future Work}
While our dual GNN architecture demonstrates significant improvements in handshape recognition, some challenges remain.  Firstly, our sequences lack precise, frame-level handshape annotations, which forced us to use stability-based heuristics to identify the most informative frames.  Other heuristics could work much better and should be explored.  We also only tested the approach on one data set, due to the novelty of the task.  Similarly augmenting other isolated sign datasets with handshape information will provide additional datasets for improved model benchmarking.  However, sequences involving multiple signs could present a unique challenge that motivates changes in our architecture.  

Our work also opens several promising directions.~First, integrating our handshape recognition system into translation models could refine phonological distinctions, as our results suggest that explicit handshape modeling improves classification performance. Our framework could also aid linguists in analyzing handshape patterns across sign languages.




\bibliography{custom}

\onecolumn  

\section*{Appendix A:}

\subsection*{A.1 Unsupervised Clustering Analysis}
To validate the stability of our unsupervised clustering analysis, we compared cluster assignments across different k values using both cosine and Euclidean distance metrics. Figures 6 and 7 show the stability comparison between k=30 and k=50 configurations.

For each cluster i in the 30-cluster configuration and cluster j in the 50-cluster configuration, we compute the Jaccard similarity coefficient:

\[ \text{Stability}(i,j) = \frac{|C_i \cap C_j|}{|C_i \cup C_j|} \]

where $C_i$ and $C_j$ are the sets of frames assigned to clusters i and j respectively. The resulting matrices visualize the overlap between clusters, with higher values (darker colors) indicating stronger correspondence between cluster assignments.

Both distance metrics show similar stability patterns, with consistent block-diagonal structures indicating that many cluster assignments are preserved even when increasing k. The relatively uniform distribution of similarity scores (predominantly in the 0.01-0.02 range) suggests that clusters maintain coherent substructure when split into finer groupings.

This supports our observation that handshape patterns emerge consistently across different clustering granularities, providing evidence for natural geometric groupings in the data.

\begin{figure}[h]
\centering
\includegraphics[width=0.9\textwidth]{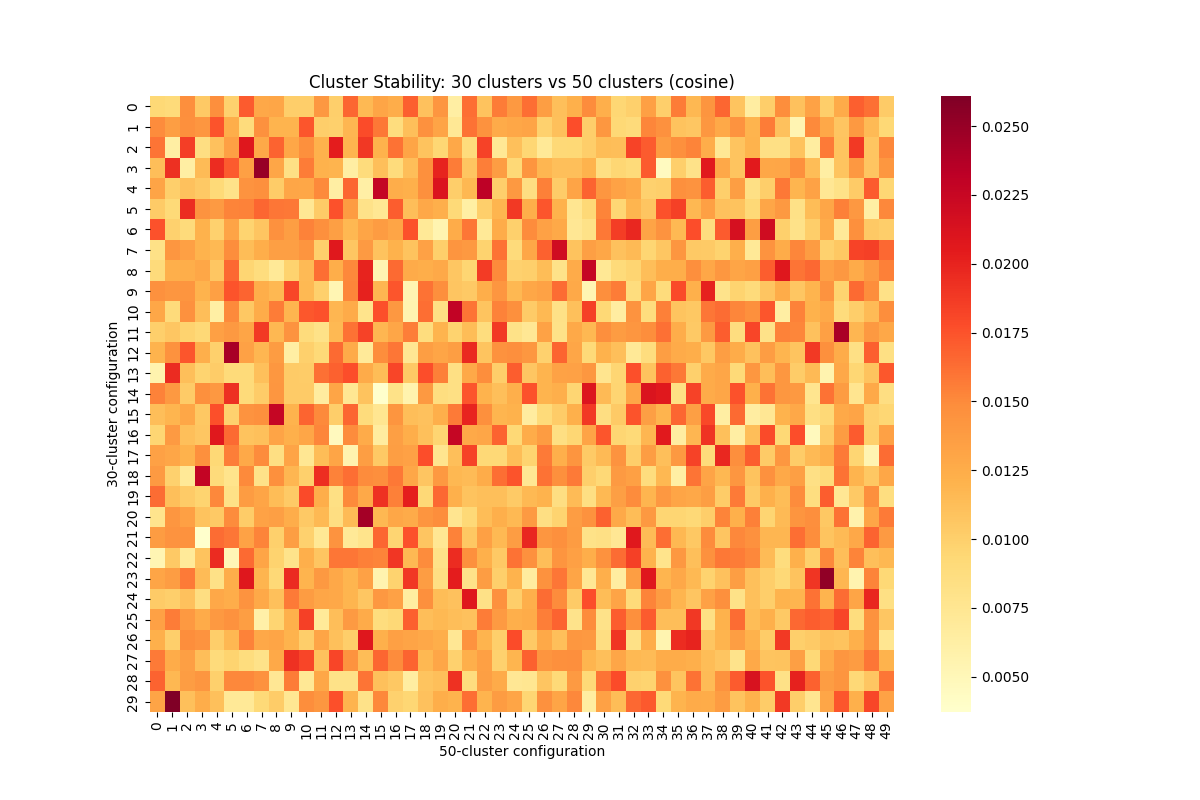}
\caption{Cluster stability comparison between 30-cluster and 50-cluster configurations using cosine distance. Each cell (i,j) shows the Jaccard similarity between cluster i from the 30-cluster configuration (y-axis) and cluster j from the 50-cluster configuration (x-axis). Darker colors indicate higher overlap between cluster assignments.}
\label{fig:stability_cosine}
\end{figure}

\begin{figure}[h]
\centering
\includegraphics[width=0.9\textwidth]{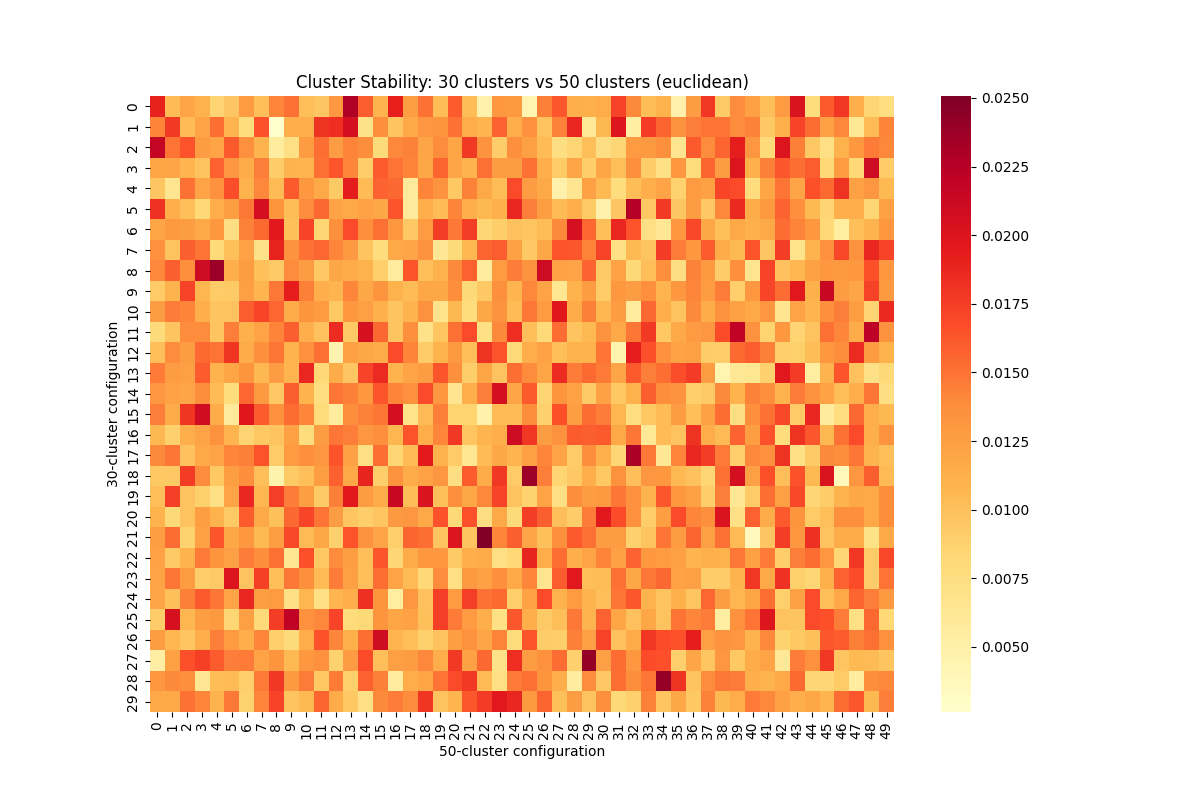}
\caption{Cluster stability comparison between 30-cluster and 50-cluster configurations using Euclidean distance. The heatmap visualizes Jaccard similarities between clusters, demonstrating consistent pattern preservation across different k values.}
\label{fig:stability_euclidean}
\end{figure}

\clearpage  

\subsection*{A.2 Detailed Model Performance}

Table A1 presents the complete per-class performance metrics for our dual GNN architecture on its last epoch. The results demonstrate the model's varying effectiveness across different handshape classes, with particularly strong performance on distinctive configurations like 'ily' (F1: 0.727) and 'w' (F1: 0.723).

\begin{longtable}{lrrrr}
\caption{Per-Class Model Performance Metrics} \label{tab:model-performance} \\
\toprule
Handshape & Precision & Recall & F1-Score & Support \\
\midrule
\endfirsthead

\multicolumn{5}{c}{\textit{Table continued from previous page}} \\
\toprule
Handshape & Precision & Recall & F1-Score & Support \\
\midrule
\endhead

\midrule
\multicolumn{5}{r}{\textit{Continued on next page}} \\
\endfoot

\bottomrule
\endlastfoot

ily & 0.857 & 0.632 & 0.727 & 19 \\
w & 0.773 & 0.680 & 0.723 & 25 \\
y & 0.644 & 0.532 & 0.583 & 109 \\
v & 0.491 & 0.596 & 0.538 & 89 \\
c & 0.617 & 0.467 & 0.532 & 107 \\
flat\_4 & 0.500 & 0.563 & 0.529 & 16 \\
a & 0.506 & 0.534 & 0.520 & 223 \\
i & 0.778 & 0.389 & 0.519 & 18 \\
spread\_open\_e & 0.531 & 0.500 & 0.515 & 34 \\
f & 0.570 & 0.464 & 0.511 & 97 \\
open\_b & 0.387 & 0.680 & 0.494 & 416 \\
s & 0.410 & 0.616 & 0.492 & 203 \\
curved\_v & 0.615 & 0.400 & 0.485 & 20 \\
flat\_h & 0.583 & 0.396 & 0.472 & 53 \\
h & 0.532 & 0.424 & 0.472 & 59 \\
open\_h & 0.444 & 0.500 & 0.471 & 16 \\
1 & 0.402 & 0.555 & 0.466 & 364 \\
o & 0.451 & 0.477 & 0.464 & 107 \\
flat\_b & 0.702 & 0.324 & 0.443 & 102 \\
4 & 0.571 & 0.353 & 0.436 & 34 \\
bent\_1 & 0.447 & 0.422 & 0.434 & 90 \\
g & 0.500 & 0.367 & 0.424 & 49 \\
p & 0.522 & 0.353 & 0.421 & 34 \\
flat\_o & 0.596 & 0.322 & 0.418 & 87 \\
open\_8 & 0.577 & 0.326 & 0.417 & 46 \\
curved\_5 & 0.618 & 0.313 & 0.416 & 67 \\
closed\_b & 0.427 & 0.376 & 0.400 & 85 \\
5 & 0.399 & 0.390 & 0.395 & 228 \\
8 & 0.688 & 0.244 & 0.361 & 45 \\
3 & 0.636 & 0.219 & 0.326 & 32 \\
l & 0.517 & 0.221 & 0.309 & 68 \\
flatspread\_5 & 0.567 & 0.213 & 0.309 & 80 \\
flat\_1 & 0.696 & 0.188 & 0.296 & 85 \\
baby\_o & 0.383 & 0.240 & 0.295 & 129 \\
curved\_l & 0.417 & 0.227 & 0.294 & 22 \\
d & 0.667 & 0.118 & 0.200 & 17 \\
bent\_v & 0.200 & 0.182 & 0.191 & 11 \\
\midrule
\multicolumn{4}{l}{Macro Avg F1: 0.440} & \\
\multicolumn{4}{l}{Accuracy: 46.07\%} & \\
\end{longtable}

\clearpage
\subsection*{A.3 Biomechanical Metrics Analysis}

The distribution patterns of our biomechanical metrics reveal several key characteristics of handshape production in ASL. Figures A3-A5 visualize these distributions.

\begin{figure}[h]
\centering
\includegraphics[width=0.9\textwidth]{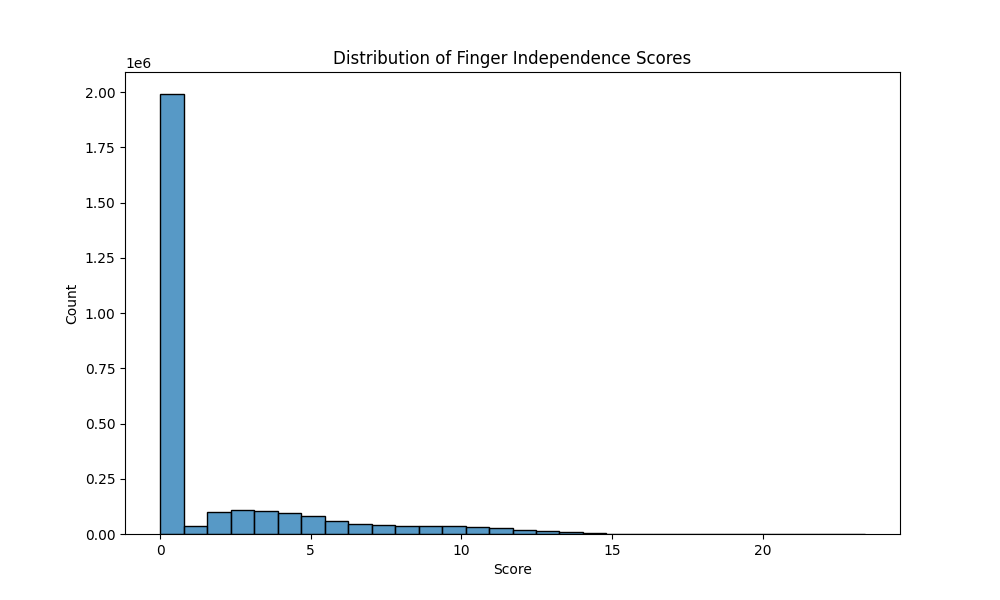}
\caption{Distribution of Finger Independence Scores across handshapes. The strongly right-skewed distribution with a peak near zero indicates that ASL handshapes predominantly favor coordinated finger movements, with relatively few configurations requiring high degrees of independent finger articulation.}
\label{fig:finger_independence}
\end{figure}

\begin{figure}[h]
\centering
\includegraphics[width=0.9\textwidth]{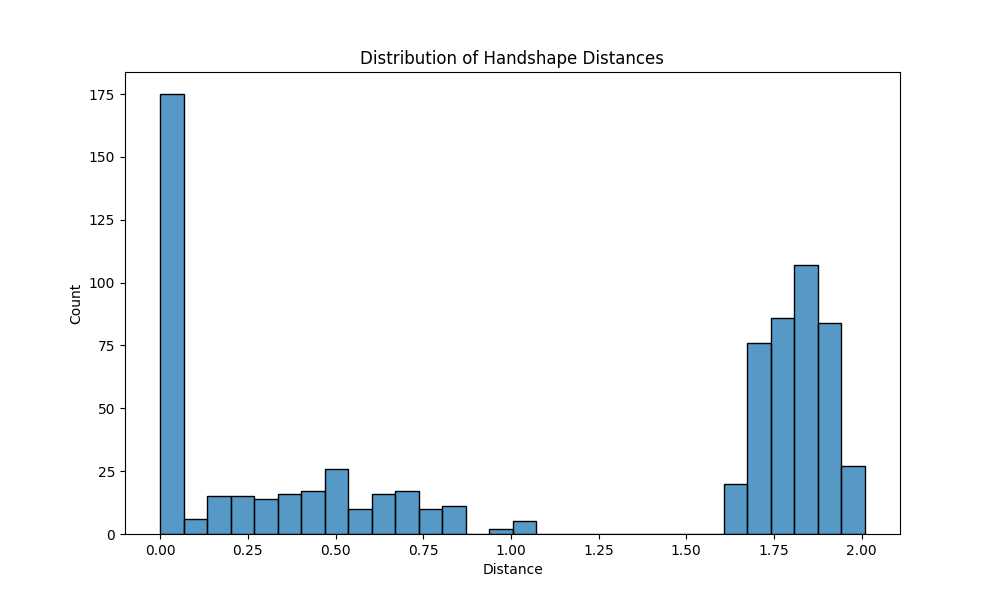}
\caption{Distribution of Handshape Distances showing a bimodal pattern. The peaks at approximately 0.0 and 1.75-2.0 suggest that handshapes in the dataset tend to be either very similar to each other or markedly different, with relatively few instances of intermediate similarity.}
\label{fig:handshape_distances}
\end{figure}

\begin{figure}[h]
\centering
\includegraphics[width=0.9\textwidth]{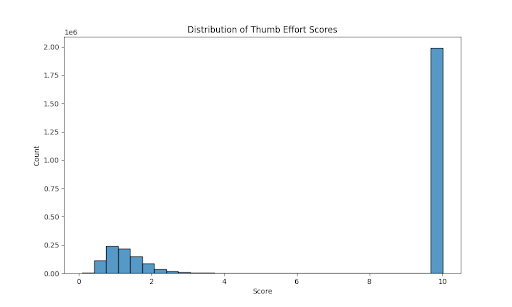}
\caption{Distribution of Thumb Effort Scores showing a clear bimodal pattern. Most handshapes exhibit relatively low thumb effort scores (0-2), while a distinct spike at score 10 indicates a subset of handshapes requiring significant thumb engagement, suggesting natural categorization in thumb usage.}
\label{fig:thumb_effort}
\end{figure}

These distributions provide quantitative evidence for biomechanical constraints in handshape production, with clear patterns in how fingers are coordinated, how the thumb is employed, and how handshapes are distinguished from each other in the signing space.

\clearpage
\subsection*{A.4 Dataset Characteristics}
The integration of ASL-LEX data revealed systematic patterns in handshape usage across 37 unique handshapes. One-handed signs (23,508 instances) showed greater variety in handshape usage than symmetrical signs (7,744 instances), which exhibited more restricted distributions due to biomechanical constraints, aligning with phonological principles of symmetry in sign language (Brentari \& Eccarius, 2010).

\end{document}